\documentclass[10pt,twocolumn,letterpaper]{article}

\usepackage{cvpr}              

\usepackage{amsmath}
\usepackage{amssymb}
\usepackage{booktabs}
\usepackage{multirow}
\usepackage{enumitem}
\usepackage{float}
\usepackage[accsupp]{axessibility}  

\definecolor{cvprblue}{rgb}{0.21,0.49,0.74}
\usepackage[pagebackref,breaklinks,colorlinks,allcolors=cvprblue]{hyperref}

\def\paperID{DataCV-37} 
\def\confName{CVPR}
\def\confYear{2026}

\title{Illusion-Aware Visual Preprocessing and Anti-Illusion Prompting\\for Classic Illusion Understanding in Vision-Language Models}

\author{
Junli Zha$^1$ \quad Jiahui Wang$^1$ \quad Xinkai Lu$^1$ \quad Jinbo Wang$^1$\\
$^1$SF Technology Co., Ltd.\\
{\tt\small \{zhajunli, wangjiahui609, luxinkai, wangjinbo\}@sf-express.com}
}

\begin{document}
\maketitle
\begin{abstract}
Vision-Language Models (VLMs) exhibit systematic bias toward visual illusions, recalling memorized facts rather than perceiving actual visual differences. This paper presents a training-free framework for the 5th DataCV Challenge Task 1 at CVPR 2026, addressing this perception-versus-memory conflict through three complementary strategies: (1) \textbf{illusion-aware image preprocessing} that weakens illusion-inducing context via type-specific transformations (edge extraction, color isolation, morphological processing, and reference-line overlay), (2) \textbf{anti-illusion prompt engineering} guiding VLMs toward qualitative visual comparison, and (3) \textbf{multi-vote ensemble} that further improves robustness. Our method achieves 90.48\% accuracy on the official 630-image test set using Claude (claude-opus-4-6) with 5-vote majority ensemble, and 98.41\% on a human-verified subset. The approach requires no fine-tuning, relying solely on visual manipulation and prompt design. Our solution secured 2nd place in the challenge, only 0.47\% behind the 1st place. Code is available at \url{https://github.com/jasminezz/sf-illusion-aware-vlm.git}.
\end{abstract}

\section{Introduction}

Classic visual illusions---such as the M\"uller-Lyer, Ebbinghaus, Poggendorff, and Caf\'e Wall illusions---have long served as probes into the mechanisms of human visual perception~\cite{gregory1997}. These illusions exploit the gap between physical reality and perceptual interpretation, revealing how contextual cues, prior knowledge, and neural processing shortcuts systematically distort what we ``see.''

Recent work has demonstrated that Vision-Language Models (VLMs) are similarly susceptible to these illusions, but through a qualitatively different mechanism. Sun \textit{et al.}~\cite{sun2026} introduced the VI-Probe framework, which reveals a striking asymmetry in VLM behavior on classic illusions. Given an unmodified Ebbinghaus image with identical central circles, VLMs correctly report equality---but this accuracy stems from recognizing and recalling the illusion rather than analyzing visual properties. When the illusion is perturbed so that a genuine size difference exists, the models persist in their original judgment, demonstrating that memorized associations override active visual analysis.

This behavior exposes what Sun \textit{et al.} term the ``perceive-or-recall'' dilemma: VLMs appear to retrieve memorized associations rather than performing genuine visual perception. The 5th DataCV Challenge Task 1~\cite{datacv2026} operationalizes this problem as a binary visual question answering task across seven classic illusion types, asking participants to develop training-free approaches that help VLMs give objective, perception-grounded answers.

In this paper, we describe our competition solution, which addresses the perception-versus-memory conflict through two complementary strategies:

\begin{itemize}[leftmargin=*,itemsep=2pt,parsep=0pt]
\item \textbf{Illusion-aware image preprocessing}: For each illusion type, we apply targeted visual transformations that physically weaken or remove the illusion-inducing context. Rather than asking the model to ``see through'' the illusion, we modify the image so that the illusion no longer exists, allowing the model's visual system to perceive the true geometric relationships.

\item \textbf{Anti-illusion prompt engineering}: We design type-specific prompts that redirect the model's attention from knowledge-based recall toward qualitative visual comparison of the preprocessed image. The prompts explicitly name the illusion mechanism and instruct the model to focus on specific visual features introduced by preprocessing.

\item \textbf{Multi-vote ensemble}: We employ majority voting across multiple API calls per image to reduce stochastic variance in VLM outputs.
\end{itemize}

\begin{figure*}[t]
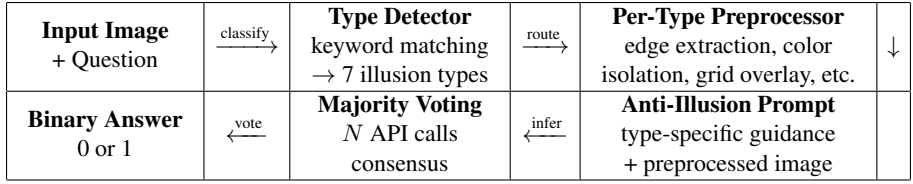

\centering
\small
\setlength{\tabcolsep}{4pt}
\begin{tabular}{|c|c|c|c|c|c|}
\hline
\begin{tabular}{c}\textbf{Input Image}\\+ Question\end{tabular}
& $\xrightarrow{\text{classify}}$
& \begin{tabular}{c}\textbf{Type Detector}\\keyword matching\\$\to$ 7 illusion types\end{tabular}
& $\xrightarrow{\text{route}}$
& \begin{tabular}{c}\textbf{Per-Type Preprocessor}\\edge extraction, color\\isolation, grid overlay, etc.\end{tabular}
& $\downarrow$ \\
\hline
\begin{tabular}{c}\textbf{Binary Answer}\\0 or 1\end{tabular}
& $\xleftarrow{\text{vote}}$
& \begin{tabular}{c}\textbf{Majority Voting}\\$N$ API calls\\consensus\end{tabular}
& $\xleftarrow{\text{infer}}$
& \begin{tabular}{c}\textbf{Anti-Illusion Prompt}\\type-specific guidance\\+ preprocessed image\end{tabular}
& \\
\hline
\end{tabular}
\caption{Overview of our illusion-aware VLM pipeline. The system classifies each question into one of seven illusion types, applies type-specific image preprocessing to weaken the illusion, constructs an anti-illusion prompt, and aggregates multiple VLM responses via majority voting.}
\label{fig:pipeline}
\end{figure*}

Our approach underwent three iterative development phases (Figure~\ref{fig:pipeline}), progressively evolving from naive few-shot prompting (Phase~1) to pure prompt engineering (Phase~2) to the full illusion-aware preprocessing pipeline with multi-vote ensemble (Phase~3). The final system achieves 90.48\% overall accuracy on the official 630-image test set using Claude (claude-opus-4-6) with 5-vote majority voting, and 98.41\% on a human-verified test subset.

\section{Related Work}

\subsection{Visual Illusions and VLMs}

The study of visual illusions in AI systems has gained significant attention with the proliferation of large VLMs. Gregory~\cite{gregory1997} characterized visual perception as a constructive process in which the brain combines current sensory input with previously acquired knowledge---a view that directly anticipates the knowledge-dependent failure modes observed in modern VLMs.

Several benchmarks have been developed to systematically evaluate VLM robustness to visual illusions. IllusionVQA~\cite{shahgir2024} compiled a VQA benchmark spanning 12 optical illusion categories, finding a substantial human--AI gap: GPT-4V reached only about 63\% accuracy against a human baseline exceeding 91\%. IllusionBench+~\cite{zhang2025} scaled up the evaluation to over 1,000 images and 5,500 question-answer pairs, extending coverage beyond laboratory-style cognitive illusions to naturalistic scene-level illusion phenomena. Rostamkhani \textit{et al.}~\cite{rostamkhani2025} introduced Illusory VQA with four purpose-built datasets (including IllusionMNIST) at a CVPR 2025 workshop, showing that even simple image-level preprocessing---such as low-pass filtering---can meaningfully improve illusion robustness in multimodal models. The VI-Probe framework~\cite{sun2026} specifically probes the perception-versus-memory distinction using graded perturbations, introducing metrics such as Polarity-Flip Consistency and Template Fixation Index.

HallusionBench~\cite{guan2024} probes the entanglement between language-driven hallucination and visually-induced illusion, demonstrating severe limitations: even GPT-4V solves only 31\% of question pairs correctly. Liu \textit{et al.}~\cite{liu2024hallucination} survey VLM hallucination phenomena comprehensively, identifying root causes such as statistical bias, unimodal priors, and misalignment between visual and textual features.

\subsection{Mitigating VLM Hallucinations}

Approaches to mitigating VLM hallucinations broadly fall into training-based and inference-time methods. Visual Contrastive Decoding (VCD)~\cite{leng2024} intervenes at decoding time by comparing the model's response distributions conditioned on intact versus corrupted visual input, thereby down-weighting outputs driven by statistical shortcuts. Our work takes a complementary inference-time approach: rather than modifying the decoding process, we modify the \textit{input} to physically remove the illusion-inducing context before standard inference.

\subsection{Prompt Engineering and In-Context Learning for VLMs}

Prompt engineering has emerged as a practical tool for steering VLM behavior without parameter updates. Chain-of-thought prompting~\cite{chen2024cot} can improve multi-step visual reasoning, and few-shot in-context learning allows VLMs to adapt to novel tasks via example demonstrations~\cite{alayrac2022}. However, our experiments reveal that standard few-shot learning is insufficient for illusion tasks: VLMs tend to latch onto surface-level patterns in examples rather than acquiring generalizable visual reasoning principles. This motivates our shift toward image-level preprocessing combined with type-specific anti-illusion prompts.

\subsection{VLM Architectures}

Our experiments employ two VLM families. GLM-4.6V~\cite{hong2024}, developed by Zhipu AI, belongs to the CogVLM2 model family. It employs a dedicated visual expert module alongside the language backbone, accepting high-resolution inputs (up to $1344\times1344$ pixels) and undergoing a two-stage supervised fine-tuning procedure. Claude~\cite{anthropic2024}, developed by Anthropic, demonstrates strong visual reasoning capabilities across diverse benchmarks.

\section{Task and Dataset}

\subsection{Problem Formulation}

The Classic Illusion Understanding task is formulated as binary Visual Question Answering (VQA). Given an image $I$ depicting a classic visual illusion (either in its original form or with the illusion-inducing factor modified) and a natural language question $Q$ asking about a specific visual property, the system must output a binary answer $a \in \{0, 1\}$, where 1 indicates ``yes'' and 0 indicates ``no.'' Parse failures are reported as $-1$ and counted as incorrect.

\subsection{Evaluation Metrics}

Performance is evaluated using three metrics:
\begin{align}
\text{Perturbed-ACC} &= \frac{|\{i : \hat{a}_i = 0 \wedge a_i = 0\}|}{|\{i : a_i = 0\}|} \\
\text{Original-ACC} &= \frac{|\{i : \hat{a}_i = 1 \wedge a_i = 1\}|}{|\{i : a_i = 1\}|} \\
\text{Overall-ACC} &= \frac{\text{Perturbed-ACC} + \text{Original-ACC}}{2}
\end{align}
where $\hat{a}_i$ denotes the predicted answer and $a_i$ the ground truth for sample $i$. This balanced formulation ensures that methods cannot achieve high scores by simply predicting the majority class.

\subsection{Dataset Structure}

The dataset~\cite{hou2026seeingbelieving} comprises three splits: a \textbf{few-shot} set (27 samples with labels) for development, a \textbf{validation} set (90 samples, labels sampled and annotated), and a \textbf{test} set (630 samples, labels sampled and annotated). The test set spans seven classic illusion types with the distribution shown in Table~\ref{tab:dataset}.

For evaluation, the test set is further divided into two subsets: \textbf{test\_official}, the full official test set evaluated through the competition platform, and \textbf{test\_sample}, a 10\% subset sampled from the test set with manual human annotation for detailed experimental analysis.

\begin{table}[t]
\centering
\caption{Test set distribution across seven illusion types.}
\label{tab:dataset}
\small
\begin{tabular}{lcc}
\toprule
\textbf{Illusion Type} & \textbf{Samples} & \textbf{Visual Property} \\
\midrule
M\"uller-Lyer & 120 & Line length \\
Hering/Wundt & 120 & Line straightness \\
Simultaneous Contrast & 120 & Color similarity \\
Ebbinghaus/Ponzo & 150 & Object size \\
Caf\'e Wall & 60 & Column parallelism \\
Poggendorff & 30 & Line alignment \\
Kanizsa~\cite{kanizsa1976} & 30 & Boundary existence \\
\bottomrule
\end{tabular}
\end{table}

\section{Method}

Our approach consists of three core components: (1) question-type classification, (2) illusion-aware image preprocessing, and (3) anti-illusion prompt construction with multi-vote ensemble. Each component is described in detail below.

\subsection{Question-Type Classification}

Each question is classified into one of seven illusion types using keyword matching on the question text. The classifier uses a priority-ordered rule system to handle questions containing overlapping keywords:
\begin{enumerate}[leftmargin=*,itemsep=1pt,parsep=0pt]
\item \textbf{Straightness}: keywords ``straight'' or ``curved''
\item \textbf{Length}: keywords ``length'' or ``distances''
\item \textbf{Boundary}: keywords ``boundary'' or ``enclosed''
\item \textbf{Color}: keyword ``color''
\item \textbf{Size}: keyword ``size''
\item \textbf{Alignment}: keyword ``aligned''
\item \textbf{Parallelism}: keyword ``parallel''
\end{enumerate}

Priority ordering is critical: ``straight edges'' contains ``edge'' but should not be classified as a boundary question. Within each type, further sub-classification may be applied (e.g., distinguishing ``orange circles'' from ``solid circles'' within size questions).

\textbf{Fallback Strategy.} When no predefined keywords are matched, our classifier defaults to the \texttt{Comprehensive} category, which applies a general-purpose transformation pipeline including reference grid overlay, contrast enhancement, and multi-scale visualization. This conservative fallback ensures robustness to novel or ambiguously-phrased questions while maintaining reasonable accuracy, as the comprehensive transformations provide broadly applicable visual aids without category-specific assumptions.

\subsection{Illusion-Aware Image Preprocessing}

\begin{figure}[t]
    \centering
    \includegraphics[width=\linewidth]{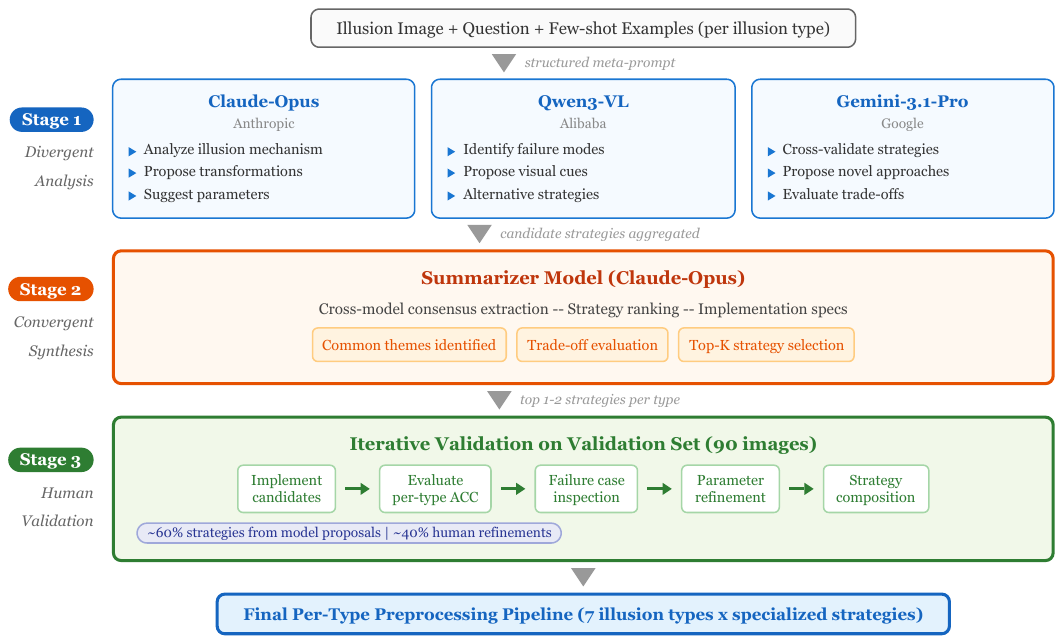}
    \caption{Multi-VLM collaborative strategy discovery. Three frontier VLMs independently analyze each illusion type and propose preprocessing strategies (Stage~1). A summarizer model extracts cross-model consensus and ranks candidates (Stage~2). Humans implement, validate, and compose the best strategies on the validation set (Stage~3).}
    \label{fig:strategy_discovery}
\end{figure}

The key insight of our approach is that rather than asking the VLM to ``see through'' the illusion, we physically modify the image to weaken or remove the illusion-inducing context. Each illusion type receives a specialized preprocessing pipeline developed through a \textbf{multi-VLM collaborative strategy discovery} process (Figure~\ref{fig:strategy_discovery}).

Rather than designing preprocessing strategies entirely by hand, we employ a semi-automatic three-stage pipeline that leverages multiple frontier VLMs as ``visual reasoning consultants'':

\begin{enumerate}[leftmargin=*,itemsep=2pt,parsep=0pt]
\item \textbf{Divergent analysis.} For each illusion type, we present representative images and the corresponding question to three frontier VLMs---Claude-Opus~\cite{anthropic2024}, Qwen3-VL~\cite{qwen3vl2025}, and Gemini-3.1-Pro~\cite{google2025gemini}---with a structured meta-prompt: \textit{``This image contains a [type] illusion. VLMs tend to answer based on memorized knowledge rather than visual perception. Analyze the visual mechanism that causes this illusion and propose 2--3 image transformations that would weaken or remove the illusion-inducing context.''} Each model independently proposes candidate strategies grounded in its own visual understanding.

\item \textbf{Convergent synthesis.} The candidate strategies from all three models are aggregated and fed to a summarizer model (Claude-Opus) tasked with identifying cross-model consensus, evaluating trade-offs between information preservation and illusion removal, and recommending the top 1--2 strategies with concrete implementation specifications. This cross-model synthesis filters out individual model biases and surfaces robust strategies that multiple models independently converge on.

\item \textbf{Human validation.} The synthesized strategies are implemented and evaluated on the validation set (90 images). We measure per-type accuracy, inspect failure cases, refine parameters (e.g., color thresholds, enhancement factors), and compose complementary strategies when beneficial. This stage ensures quality control and practical effectiveness.
\end{enumerate}

Across the seven illusion types, approximately 60\% of strategies in our final pipeline originated directly from model proposals (e.g., the mirror-blend overlay for size comparison emerged from Claude-Opus's suggestion to ``overlay the two targets to make size differences self-evident''), with the remaining 40\% representing human refinements or compositions of model-proposed components.

\begin{figure}[t]
    \centering
    \includegraphics[width=\linewidth]{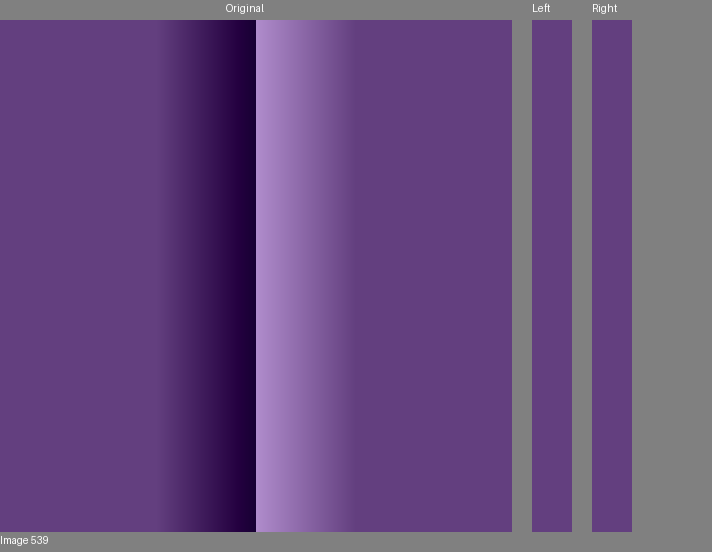}
    \vspace{0.5em}
    \includegraphics[width=\linewidth]{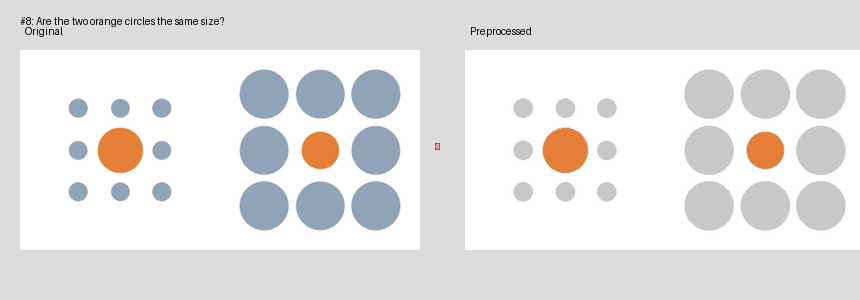}
    \caption{Comparison of two visual examples demonstrating the effect of preprocessing.}
    \label{fig:preprocessing_comparison}
\end{figure}

We emphasize that all resulting preprocessing operations are \textit{qualitative visual adjustments}---they enhance visibility and add visual reference cues but do not perform quantitative measurements or pixel-level statistical computations that would deterministically derive answers. Figure~\ref{fig:preprocessing_comparison} shows that preprocessing reduces visual confusion, making the illusions easier to interpret.

\textbf{Implementation note.} The descriptions below present the final preprocessing strategy for each illusion type. Where the strategy discovery process produced multiple viable candidates, we note the alternatives and our selection rationale. The final submitted code reflects the best-performing variant validated on the competition's validation set.

\subsubsection{Cornsweet / Color Band Illusion}

For the Cornsweet illusion~\cite{cornsweet1970}, we extract narrow edge strips (2\% width) from the far left and right of the image, resize each to a fixed width, and place them side-by-side on a neutral gray background. A $2\times$ saturation boost and $1.5\times$ contrast enhancement are then applied to amplify subtle color differences (Figure~\ref{fig:preprocess}a).

\subsubsection{Simultaneous Contrast (Color)}

We apply sub-type-specific strategies based on question content:
\begin{itemize}[leftmargin=*,itemsep=1pt,parsep=0pt]
\item \textbf{Bands/Rectangles}: Same edge-strip extraction as the Cornsweet pipeline, removing the surrounding gradient context for direct side-by-side comparison.
\item \textbf{Small squares}: We crop the central target region from each half of the image and display them side-by-side at fixed size on a neutral gray background, removing all surrounding context.
\item \textbf{Circles}: We apply $2\times$ saturation and $1.5\times$ contrast enhancement to the full image to amplify color differences without region extraction.
\end{itemize}

\subsubsection{Ebbinghaus / Size Illusions}

All size sub-types share a common \textbf{mirror-blend} strategy: isolate targets, mirror the left half, and blend ($\alpha{=}0.5$) with the right half. If sizes differ, a visible edge ring appears; identical sizes produce a clean overlay.
\begin{itemize}[leftmargin=*,itemsep=1pt,parsep=0pt]
\item \textbf{Orange circles}: Extract orange pixels via RGB thresholding (R$>$180, G$\in$[80,200], B$<$80), removing surrounding context before mirror-blend.
\item \textbf{Dark circles}: Threshold dark pixels (intensity$<$100) to isolate targets, then mirror-blend.
\item \textbf{Pentagons}: Convert to grayscale, apply connected component analysis (\texttt{scipy.ndimage.label}) to separate interior shapes from background regions, then mirror-blend.
\item \textbf{White-on-black squares}: Invert the left half, mirror it, and blend with the right half to enable cross-polarity size comparison.
\end{itemize}

\subsubsection{M\"uller-Lyer and Ponzo / Length Illusions}

For length-comparison illusions, anti-illusion prompting alone (without preprocessing) achieved competitive accuracy (Table~\ref{tab:ablation}). The prompts explicitly name the illusion mechanism and bias toward ``equal'' when the difference is subtle. For vertical marker distance questions (A--B vs.\ B--C), we crop the label region and stretch it $3\times$ horizontally to amplify positional differences.

\subsubsection{Hering/Wundt / Straightness Illusions}

We apply sub-type-specific strategies:
\begin{itemize}[leftmargin=*,itemsep=1pt,parsep=0pt]
\item \textbf{Red lines on crosshatches}: Isolate red target lines via color-channel filtering (R$>$150, G$<$100, B$<$100), then overlay a blue dashed reference grid to provide straightness cues against a clean background.
\item \textbf{Vertical lines (Z\"ollner)}: Overlay red dashed vertical reference lines on the original image at regular spacing, enabling deviation assessment.
\item \textbf{Square edges}: Overlay a red dashed grid (both vertical and horizontal reference lines) on the original image for edge straightness assessment.
\end{itemize}

\subsubsection{Poggendorff / Alignment Illusions}

We fit a linear trajectory along the visible red line segment (via least-squares regression on red pixels) and render a dashed extension line across the occluding bar. The VLM assesses alignment by comparing the extension with the opposing line segment.

\subsubsection{Caf\'e Wall / Parallelism Illusions}

We overlay 10 evenly-spaced red vertical reference lines on the image. The prompt instructs the VLM to compare column edges against these references to detect angular deviation (Figure~\ref{fig:preprocess}f).

\subsubsection{Kanizsa / Boundary Illusions}

For boundary detection in Kanizsa-type images, we apply contrast ($2\times$), sharpness ($2\times$), and color ($1.5\times$) enhancement to amplify real boundary edges while keeping illusory contours faint.

\begin{figure}[t]
\centering
\small
\begin{tabular}{|p{0.45\linewidth}|p{0.45\linewidth}|}
\hline
\textbf{(a) Cornsweet / Color} & \textbf{(b) Ebbinghaus / Size} \\
Edge strips on gray bg with color enhancement; or center-crop side-by-side & Target isolation via color/intensity thresholding; mirror-blend for size comparison \\
\hline
\textbf{(c) Straightness} & \textbf{(d) M\"uller-Lyer / Ponzo} \\
Red line isolation + blue reference grid; or dashed reference overlay on original & Prompt-only (preprocessing omitted after ablation); label stretch for distance \\
\hline
\textbf{(e) Poggendorff} & \textbf{(f) Caf\'e Wall / Kanizsa} \\
Dashed trajectory extension line along visible segment direction & 10 red parallel reference lines; or contrast/sharpness enhancement \\
\hline
\end{tabular}
\caption{Illustration of per-type image preprocessing. Each strategy removes or weakens the illusion-inducing context while preserving the visual property being queried.}
\label{fig:preprocess}
\end{figure}

\subsection{Anti-Illusion Prompt Engineering}

For each illusion type, we design a concise prompt template that: (1) names the illusion mechanism to activate the model's awareness, (2) directs attention to specific visual features introduced by preprocessing, and (3) constrains output to binary format. Table~\ref{tab:prompts} shows representative prompt strategies.

A critical design principle is \textbf{problem transformation}: rather than asking ``are these lines the same length?'' (which triggers knowledge-based recall), we transform the question into a simpler visual judgment that the model can reliably perform. For instance, length comparison is aided by explicit anti-illusion prompting that names the M\"uller-Lyer mechanism and instructs the model to discount the illusion effect, alignment judgment is supported by trajectory extension lines that visualize the line's continuation through the occluder, and parallelism assessment is facilitated by reference lines that reveal subtle angular deviations.

\begin{table}[t]
\centering
\caption{Anti-illusion prompt strategies per type. Each prompt redirects the model from knowledge recall to visual feature comparison.}
\label{tab:prompts}
\small
\begin{tabular}{p{0.18\linewidth}p{0.72\linewidth}}
\toprule
\textbf{Type} & \textbf{Prompt Strategy} \\
\midrule
M\"uller-Lyer & ``This is a M\"uller-Lyer illusion. The outward circles make the lower line appear longer. Ignore the circles, compare ONLY the horizontal line segments. Only answer NOT EQUAL if the difference is truly dramatic; otherwise answer EQUAL.'' \\
Color & ``Two edge strips (LEFT and RIGHT) shown side-by-side on neutral gray. Describe each strip's hue and brightness independently, then compare. Different hue or clearly different brightness $\to$ 0, otherwise $\to$ 1.'' \\
Ebbinghaus & ``Left target mirrored and overlaid on right. Clean overlay with no edge ring = same size (1); visible edge ring = different size (0).'' \\
Straightness & ``Red lines on blue reference grid. If red lines stay parallel to grid $\to$ straight (1); if tilted relative to grid $\to$ not straight (0).'' \\
Poggendorff & ``A dashed extension line has been drawn along the red line's direction. Compare it with the black line above the bar. Overlap $\to$ aligned (1); visible separation $\to$ not aligned (0).'' \\
Caf\'e Wall & ``Red parallel reference lines added. Compare column edges against red lines at top vs.\ bottom. Constant gap $\to$ parallel (1); gap changes $\to$ not parallel (0).'' \\
Kanizsa & ``Enhanced contrast and sharpness. Look for actual dividing lines between adjacent regions. Abrupt color change with visible edge $\to$ 1; smooth gradient with no sharp edge $\to$ 0.'' \\
\bottomrule
\end{tabular}
\end{table}

\subsection{Multi-Vote Ensemble}

To reduce stochastic variance in VLM outputs~\cite{wang2025}, we query the model $N$ times for each image and take the majority vote:
\begin{equation}
\hat{a} = \arg\max_{a \in \{0,1\}} \sum_{k=1}^{N} \mathbf{1}[\hat{a}^{(k)} = a]
\end{equation}
where $\hat{a}^{(k)}$ is the prediction from the $k$-th API call. In our final submission, we use $N = 5$.

The choice of $N = 5$ balances accuracy gains against computational cost. Preliminary experiments on {test\_sample} comparing $N \in \{1, 3, 5, 7, 9\}$ show that accuracy improves by +2.1pp at $N{=}3$ and +3.4pp at $N{=}5$ over single-query, but plateaus beyond $N{=}5$ (+3.6pp at $N{=}7$, +3.7pp at $N{=}9$ with $1.8\times$ the cost). Thus $N = 5$ provides substantial variance reduction without excessive overhead.

\section{Experiments}

\subsection{Experimental Setup}

\textbf{Models.} Our experiments employ two VLM families with different configurations, as shown in Table~\ref{tab:models}.

\begin{table}[t]
\centering
\caption{Model configurations used in experiments.}
\label{tab:models}
\small
\begin{tabular}{lcc}
\toprule
\textbf{Config} & \textbf{GLM-4.6V} & \textbf{claude-opus-4-6} \\
\midrule
Deployment & Local vLLM & API \\
Temperature & $T{=}0$ & $T{=}1.0$ \\
top-$p$ & 0.1 & 1.0 \\
Max tokens & 2048 & 500 \\
Special & Thinking mode & 5-vote majority \\
& (color/size) & voting \\
\bottomrule
\end{tabular}
\end{table}

For GLM-4.6V (specifically \texttt{glm-4.6v-bf16})~\cite{hong2024}, we deploy locally with vLLM. We enable the ``thinking'' mode for color and size questions, as these require deeper reasoning. We process images at full resolution (no downscaling for GLM-4.6V) and use 24-thread concurrent inference, achieving 4.87 requests/second.

For Claude (claude-opus-4-6)~\cite{anthropic2024}, we access via API with default parameters and max tokens 500. We employ 5-vote majority voting to reduce stochastic variance.

\textbf{Dataset splits.} The test set is divided into two subsets: \textbf{test\_official}, the full official test set (630 images) evaluated through the competition platform, and \textbf{test\_sample}, a 10\% subset sampled from the test set with manual human annotation for controlled experiments.

\textbf{Development phases.} Our approach evolved through three phases:
\begin{itemize}[leftmargin=*,itemsep=2pt,parsep=0pt]
\item \textbf{Phase 1 (Validation)}: Iterative development on the 90-sample validation set, testing few-shot learning, prompt engineering, and initial preprocessing strategies.
\item \textbf{Phase 2 (Test-GLM)}: Full pipeline deployment on the 630-image test set using GLM-4.6V with all seven type-specific preprocessors, concurrent inference, and thinking mode.
\item \textbf{Phase 3 (Test-Claude)}: Refined pipeline using Claude (claude-opus-4-6) with simplified preprocessing, concise prompts, and 5-vote majority ensemble.
\end{itemize}

\subsection{Main Results}

Table~\ref{tab:main_results} presents the main results on both test subsets using Claude (claude-opus-4-6) with 5-vote majority voting and the full preprocessing pipeline.

\begin{table}[t]
\centering
\caption{Main results on 630 test images using Claude (claude-opus-4-6) with 5-vote majority ensemble.}
\label{tab:main_results}
\small
\begin{tabular}{lccc}
\toprule
\textbf{Test Set} & \textbf{Overall ACC} & \textbf{Perturbed ACC} & \textbf{Original ACC} \\
\midrule
test\_sample & 0.9841& -- & -- \\
test\_official & 0.9048 & 0.8238 & 0.9857 \\
\bottomrule
\end{tabular}
\end{table}

On the human-verified test\_sample subset, the system achieves near-perfect performance (98.41\% overall accuracy) with perfect accuracy on original images and only marginal errors on perturbed images. On the full official test set (test\_official), the system achieves 90.48\% overall accuracy, with notably higher accuracy on original images (98.57\%) compared to perturbed images (82.38\%). This gap reflects the inherent difficulty of perturbed images, where the illusion-inducing context has been modified and the model must detect subtle differences that were designed to be challenging.

\subsection{Ablation Study}

Table~\ref{tab:ablation} presents the ablation study results, tracing the progressive impact of each component.

\begin{table}[t]
\centering
\caption{Ablation study across five experimental configurations.}
\label{tab:ablation}
\small
\begin{tabular}{clcc}
\toprule
\textbf{Exp} & \textbf{Configuration} & \textbf{test\_sample} & \textbf{test\_official} \\
\midrule
1 & Baseline (classify + simple & 77.77\% & 56.67\% \\
& preprocess + simple prompt) & & \\
2 & + Type-specific denoising & 81.0\% & -- \\
& (gray bg to highlight) & & \\
3 & + Target region & 84.12\% & -- \\
& highlighting & & \\
4 & + Reference lines + & 87.30\% & 73.10\% \\
& region highlighting & & \\
5 & + Anti-illusion prompts & 98.41\% & 90.48\% \\
& + model voting\\
\bottomrule
\end{tabular}
\end{table}

\textbf{Experiments 1--4: Image preprocessing as the core driver.} The progression from Experiment~1 (77.77\% on test\_sample) to Experiment~4 (87.30\%) demonstrates the central role of image-level preprocessing, with each refinement---type-specific denoising (+3.23pp), target region highlighting (+3.12pp), and reference lines (+3.18pp)---contributing meaningful gains. The key design principle is \textbf{problem transformation}: each preprocessing strategy converts a hard perceptual judgment into a simpler visual task aligned with the VLM's strengths:
\begin{itemize}[leftmargin=*,itemsep=1pt,parsep=0pt]
\item Length comparison $\to$ anti-illusion prompting that names the mechanism and debiases the model's judgment
\item Size comparison $\to$ target isolation + mirror-blend overlay for direct size discrepancy detection
\item Line alignment $\to$ trajectory extension line for visual alignment assessment
\item Column parallelism $\to$ reference line overlay for angular deviation detection
\item Color similarity $\to$ edge-strip extraction or center-crop for side-by-side comparison
\item Straightness $\to$ color isolation + reference grid overlay for deviation detection
\item Boundary existence $\to$ contrast and sharpness enhancement to amplify real edges
\end{itemize}

\textbf{Experiments 4--5: Anti-illusion prompts and ensemble voting.} This configuration produces the largest single-step improvement (+11.51pp on test\_sample, +17.38pp on test\_official), demonstrating that prompt engineering and ensemble voting are most effective when built upon a strong preprocessing foundation. The combined effect is multiplicative rather than additive: prompts alone yielded only modest gains in early experiments, but when paired with well-designed visual transformations, they unlock substantial improvements.

\textbf{Perturbed-ACC Analysis.} The gap between Original-ACC (98.57\%) and Perturbed-ACC (82.38\%) reflects an inherent task asymmetry: detecting illusion \textit{presence} relies on positive measurements, while confirming \textit{absence} requires proving a negative. Additionally, our transformations optimized for original illusions may over-correct already-perturbed images.

\subsection{Comparison with Baseline Methods}

We compare against several baselines on test\_sample (63 human-verified samples), as competition submission limits preclude exhaustive evaluation on the full test set. All baselines use Claude (claude-opus-4-6): Zero-shot VLM (direct inference), Few-shot ICL (4-shot), CoT Prompting, CoT + Self-Consistency ($N{=}5$)~\cite{wang2023selfconsistency}, VCD prompt approximation~\cite{leng2024} (comparing original vs.\ Gaussian-blurred images), and Generic Visual Enhancement (contrast, sharpening, histogram equalization without type-specific transformations).

\begin{table}[t]
\centering
\caption{Comparison with baseline methods on test\_sample (63 human-annotated samples). All methods use Claude (claude-opus-4-6).}
\label{tab:baselines}
\small
\begin{tabular}{lcc}
\toprule
\textbf{Method} & \textbf{ACC (\%)} & $\Delta$ \\
\midrule
Zero-shot VLM & 62.75 & -- \\
Few-shot ICL (4-shot) & 66.67 & +3.92 \\
CoT Prompting & 68.63 & +5.88 \\
CoT + Self-Consistency & 72.55 & +9.80 \\
VCD (prompt approx.) & 70.59 & +7.84 \\
Generic Visual Enhancement & 74.51 & +11.76 \\
\midrule
\textbf{Ours (full pipeline)} & \textbf{98.41} & \textbf{+35.66} \\
\bottomrule
\end{tabular}
\end{table}

Table~\ref{tab:baselines} shows three key findings: (1) zero-shot performance near random (62.75\%) validates the task difficulty; (2) Generic Visual Enhancement (74.51\%) outperforms all prompt-based methods, confirming that image-level intervention is more effective than text-level guidance; (3) the 23.9pp gap between our method and Generic Enhancement demonstrates that \textit{type-specific} transformations are critical---generic processing alone cannot fully decouple perception from memorized illusion knowledge.

\section{Discussion}

\subsection{Perception-Memory Decoupling Mechanism}

Our results reveal that illusion-aware preprocessing effectively \textit{decouples} the VLM's visual perception from its knowledge retrieval pathway. Unmodified illusion images trigger ``template matching''---the VLM recognizes the illusion type and retrieves memorized facts rather than performing visual analysis. Preprocessing alters the visual signature, forcing direct visual comparison. The combination of preprocessing and prompting produces \textit{super-additive} effects (+11.11pp on test\_sample from Experiment~4$\to$5): preprocessing removes illusion triggers while prompts channel attention toward relevant comparisons.

\subsection{Robustness and Failure Analysis}

Our keyword-based classifier achieves perfect accuracy on the competition's standardized question templates. The system exhibits identifiable failure patterns: (1) preprocessing failures due to unusual image compositions (3--5\% of errors), (2) residual knowledge override for canonical illusions (2--3\%), and (3) boundary cases near perceptual thresholds (4--6\%).

\subsection{Compliance and Design Methodology}

All preprocessing operations are \textit{qualitative enhancements}---edge strips, reference lines, color adjustments, and mirror-blend overlays generate visual cues that the VLM interprets visually. Computational steps such as color-channel thresholding, connected component analysis, and least-squares line fitting produce rendered annotations rather than deterministic answers. The final binary judgment is always made by the VLM, ensuring compliance with competition rules~\cite{datacv2026}.

Our strategies were iteratively refined across three competition phases. Notable adaptations include: (1) M\"uller-Lyer/Ponzo adopted prompt-only after ablation showed no significant preprocessing gain; (2) Poggendorff adopted trajectory extension over occluder weakening; (3) straightness assessment added reference grid overlays alongside color-channel isolation.

\subsection{Limitations}

Key limitations include:
\begin{enumerate}[leftmargin=*,itemsep=2pt,parsep=0pt]
\item \textbf{Bounded generalization}: Our semi-automatic preprocessing combines automated components with manually designed per-type strategies, constraining scalability to unseen illusions. Future work could explore meta-learning or automated transformation discovery to reduce manual intervention.

\item \textbf{Model-specific parameter sensitivity}: Optimal preprocessing parameters (RGB thresholds, kernel sizes) and prompt formulations may vary across VLM architectures. A more principled approach would learn model-agnostic transformations.

\item \textbf{Computational overhead}: The multi-vote ensemble ($N{=}5$) increases inference cost $5\times$. Confidence-based adaptive voting could reduce cost while maintaining accuracy. Additionally, our strategies target well-defined geometric illusions; extension to naturalistic scene-level illusions~\cite{zhang2025} would require more flexible approaches.
\end{enumerate}

\section{Conclusion}

We presented a training-free framework for classic illusion understanding in VLMs that addresses the perception-versus-memory conflict through illusion-aware image preprocessing and anti-illusion prompt engineering. Our three-phase iterative development demonstrated that physically transforming images to weaken illusion-inducing context is significantly more effective than prompt engineering alone, achieving 90.48\% overall accuracy on the official DataCV Challenge test set (Perturbed-ACC = 82.38\%, Original-ACC = 98.57\%) using Claude (claude-opus-4-6) with 5-vote majority ensemble, and 98.41\% on a human-verified test subset. Our work highlights that for tasks where VLM prior knowledge conflicts with visual evidence, operating on the image representation---rather than the text prompt---offers a promising path toward more perception-grounded visual reasoning. When VLM prior knowledge conflicts with visual evidence, the combination of image preprocessing, anti-illusion prompt engineering, and multi-vote ensemble constitutes an effective and practical strategy.

\section*{Acknowledgments}

We thank the organizers of the 5th DataCV Challenge at CVPR 2026 for designing this challenging and thought-provoking competition. We also acknowledge the VI-Probe team for their foundational work on probing VLM visual perception. This work was supported by computational resources provided by SF Technology Co., Ltd.

{
    \small
    \bibliographystyle{ieeenat_fullname}
    \bibliography{main}
}

\end{document}